\documentclass[10pt,twocolumn,letterpaper]{article}

\usepackage{iccv}
\usepackage{times}
\usepackage{epsfig}
\usepackage{graphicx}
\usepackage{amsmath}
\usepackage{amssymb}

\usepackage{booktabs}
\usepackage{subcaption}
\usepackage{siunitx}
\usepackage{multirow}
\usepackage[breaklinks=true,bookmarks=false]{hyperref}

\iccvfinalcopy % *** Uncomment this line for the final submission

\def\iccvPaperID{****} % *** Enter the ICCV Paper ID here

\ificcvfinal\fi
\begin{document}
\begin{minipage}{\textwidth}
	\large
	\textbf{IEEE Copyright Notice} \\
	\\
	$\copyright$
	2019 IEEE. Personal use of this material is permitted. Permission from IEEE must be obtained
	for  all  other  uses,  in  any  current  or  future  media,  including  reprinting/republishing  this
	material for advertising or promotional purposes, creating new collective works, for resale
	or redistribution to servers or lists, or reuse of any copyrighted component of this work in
	other works.\\
	\\
	Pre-print of article that will appear at the
	\textbf{2019 International Conference on Computer Vision (ICCV 2019) - 2nd Workshop on Deep Learning for Visual SLAM.}\\

%	\textbf{BibTex:}\\
%	@inproceedings
%	$\{$\\
%	bui19iccvw,\\
%	author =
%	$\{$
%	Mai Bui and Christoph Baur and Nassir Navab and Slobodan Ilic and Shadi Albarqouni
%	$\}$
%	,\\
%	booktitle =
%	$\{$
%	International Conference on Computer Vision Workshops (ICCVW 2019)
%	$\}$
%	, \\
%	organization =
%	$\{$
%	IEEE
%	$\}$
%	,\\
%	title =
%	$\{$
%	Adversarial Networks for Camera Pose Regression and Refinement
%	$\}$
%	,\\
%	year =
%	$\{$
%	2019
%	$\}$\\
%	$\}$
%	
	
\end{minipage}

\newpage
%%%%%%%%% TITLE
\title{Adversarial Networks for Camera Pose Regression and Refinement}
%\author{Mai Bui	\thanks{ Technical University Munich, Germany} \and Christoph Baur\footnotemark[1] \and Nassir Navab\footnotemark[1] \and Slobodan Ilic \thanks{ Siemens AG, Munich, Germany} \and Shadi Albarqouni \footnotemark[1]  \footnotemark[2]}
\renewcommand\footnotemark{}
\renewcommand\footnoterule{}
\author{Mai Bui$^{1,*}$, Christoph Baur$^{1,*}$, Nassir Navab$^{1,3}$, Slobodan Ilic$^{1,2,^\dagger}$ and Shadi Albarqouni$^{1,^\dagger}$\\% <-this % stops a space
	%\thanks{*This work was not supported by any organization}% <-this % stops a space
	\thanks{$^\dagger$ S. Ilic and S. Albarqouni are joint senior authors}
	\thanks{$^*$ M. Bui and C. Baur contributed equally to this work}
	$^{1}$ Technical University of Munich, Germany\\%
	%        University of Twente, 7500 AE Enschede, The Netherlands
	%        {\tt\small albert.author@papercept.net}}%
	$^{2}$ Siemens AG, Munich, Germany\\
	$^{3}$ Johns Hopkins University, Baltimore, USA
	%        Dayton, OH 45435, USA
	%        {\tt\small b.d.researcher@ieee.org}}%
}
%\author{Mai Bui\\
%Technical University Munich\\
%\and Christoph Baur
%\and Nassir Navab
%\and
%Slobodan Ilic\\
%Siemens AG\\
%\and Shadi Albarqouni
%}

\maketitle

\definecolor{green}{RGB}{64, 138, 89}
\definecolor{blue}{RGB}{40, 40, 91}
%%%%%%%%% ABSTRACT
\begin{abstract}
	Despite recent advances on the topic of direct camera pose regression using neural networks, accurately estimating the camera pose of a single RGB image still remains a challenging task. To address this problem, we introduce a novel framework based, in its core, on the idea of implicitly learning the joint distribution of RGB images and their corresponding camera poses using a discriminator network and adversarial learning. Our method allows not only to regress the camera pose from a single image, however, also offers a solely RGB-based solution for camera pose refinement using the discriminator network. Further, we show that our method can effectively be used to optimize the predicted camera poses and thus improve the localization accuracy. To this end, we validate our proposed method on the publicly available 7-Scenes dataset improving upon the results of direct camera pose regression methods.
\end{abstract}	
\section{Introduction}
\begin{figure}[t]
	\vspace{1.1cm}
	\begin{center}
		\includegraphics[width=0.47\textwidth]{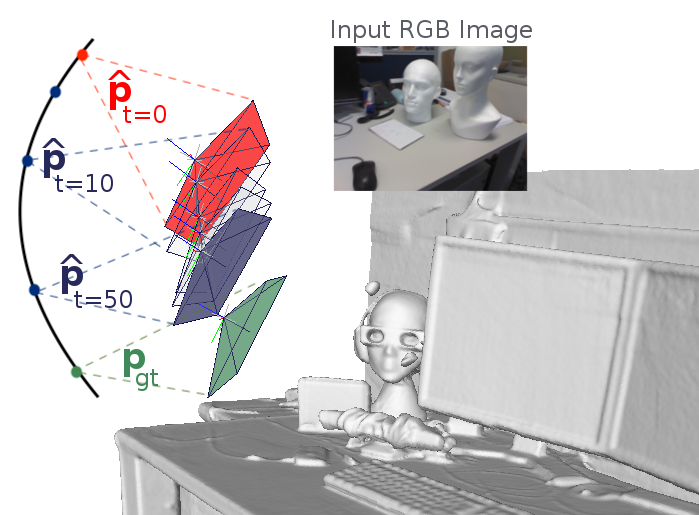}
	\end{center}
	\caption{Given an RGB input image, our method regresses a camera pose estimate $\hat{\textbf{p}}_{t=0}$ (\textcolor{red}{red}) in reference to a known scene. By incorporating adversarial training and following pose refinement, the regressed pose is updated and pushed further towards the ground truth pose $\textbf{p}_{gt}$ (\textcolor{green}{green}), resulting in the final prediction $\hat{\textbf{p}}_{t=50}$ (\textcolor{blue}{blue}).}
	\label{fig:vis}
\end{figure}
Camera re-localization is an important topic in computer vision applications such as simultaneous localization and mapping (SLAM) \cite{mur2015orb, williams2011automatic} in case of tracking failure, augmented reality \cite{marchand2016pose} or in robotics for navigation \cite{bonin2008visual}. Current methods have focused on computing the camera pose given 2D-3D correspondences between the input image and a 3D model of the scene, in essence predicting the camera pose by solving the perspective-n-point problem. Most often correspondences are computed using for example SIFT \cite{jegou2010aggregating} features or implicitly learned using regression forests \cite{shotton2013scene,valentin2015exploiting} as well as deep learning methods \cite{brachmann2017dsac,brachmann2018learning,bui2018bmvc}. On the other hand, direct camera pose estimation approaches have been developed, that regress the camera pose using convolutional neural networks (CNNs), providing a very fast solution to solve this task and, in contrast to previous methods, solely relying on RGB information \cite{kendall2015posenet,kendall2016modelling,kendall2017geometric,walch2017image}. This advantage makes such methods easily applicable in indoor as well as outdoor scenarios without requiring a 3D model or depth information. However, despite recent advances of these methods, accurately regressing the camera pose of a corresponding RGB image still remains a difficult task, especially if very little training data is available. The performance of correspondence-based methods, in comparison, can most often be accounted to an iterative pose refinement step using RANSAC, that due to the absence of a 3D model has not yet been investigated in the context of direct camera pose regression frameworks. Therefore, in this paper we make an attempt at providing a deep learning based solution for RGB-based camera pose refinement. 

For this aim, we draw inspiration from the framework of Generative Adversarial Networks (GANs) \cite{goodfellow2014generative}, which has recently shown great success in improving the performance of deep neural networks trained for tasks such as object detection \cite{wang2017fast}, human pose estimation \cite{chen2017adversarial,yang20183d} or realistic image composition \cite{lin2018st}. Such GANs consist of two networks, a generator that captures the underlying data distribution and a discriminator that estimates the probability of a sample coming from the actual distribution or the generated one, i.e. can tell the real distribution and distribution of generated data apart. During training, the two networks are in competition with each other as the generator tries to better mimic the ground truth data distribution such that it becomes more and more difficult for the discriminator to correctly classify a sample representation. More precisely, in every training step, the generator is updated in a way such that it is more likely to fool the discriminator. 

In order to improve direct camera pose regression models, 
%we attempt to leverage additional information provided by the conditional distribution of camera poses and their corresponding RGB images, an approach which to the best of our knowledge has not yet been investigated by the community. 
%, thus this process is referred to as \emph{adversarial learning}.
%Generative adversarial networks (GANs) \cite{goodfellow2014generative} consist of two networks, a generator that captures the underlying data distribution and a discriminator that estimates the probability of a sample coming from the actual distribution or the generated one. During training, the two networks are in competition with one another as the generator tries to better mimic the ground truth data distribution, so that it becomes more and more difficult for the discriminator to correctly classify a sample representation. Since their introduction GANs have had great success and it has been shown that these models can aid in improving deeply trained networks in various tasks such as object detection \cite{wang2017fast} or human pose estimation \cite{chen2017adversarial,yang20183d}.
%In this work, we would like to investigate the effect of adversarial networks on direct camera pose regression methods. 
and to better model the connection between RGB images and their camera poses, we first follow the training procedure of GANs and combine a camera pose regression network and a pose discriminator network that learns to distinguish between accurate real and potentially erroneously regressed poses and the input RGB image. This way, we attempt to implicitly model the joint distribution between an RGB image and the corresponding camera pose capturing the geometric mapping between the two in the discriminator network. Once learned, we show how the information contained in the discriminator network can be leveraged to further refine the predicted poses during inference. To summarize our contributions, we propose a novel framework for camera pose regression, that 1) includes the effect of adversarial learning in the aforementioned frameworks and 2) introduces a solely RGB-based solution for refining the resulting camera poses %, once the discriminator has successfully learned the conditional distribution of camera poses and images, we show how the discriminator network can further be used to refine the regressed poses, 
giving an additional boost in performance. An example result of our method is shown in Figure \ref{fig:vis}, where we visualize regressed and optimized camera poses in comparison to the ground truth pose.

%Thus, we attempt the geometric information that is captured between an RGB image and the corresponding camera pose

\section{Related Work}
Methods working on the topic of camera pose estimation can mainly be divided into three groups: correspondence-based, image-retrieval-based and direct pose regression approaches.

\begin{figure*}[t]
	\begin{center}
		\includegraphics[width=0.99\textwidth]{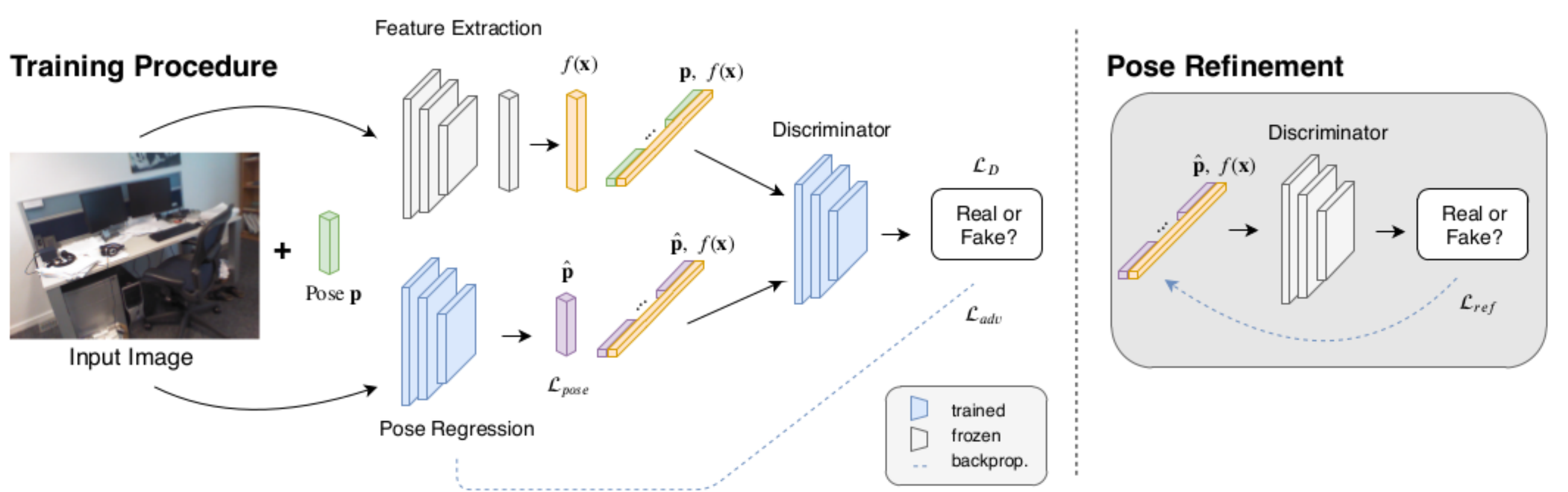}
	\end{center}
	\caption{Given an RGB image, a corresponding camera pose is estimated with a pose regression network. Alongside the estimated pose, a feature representation of the corresponding image is extracted and used to train a discriminator network. This network is trained to distinguish between ground truth and regressed poses considering the input image and can then be leveraged to refine the regressed camera pose.}
	\label{fig:framework}
\end{figure*}

\paragraph{Correspondence-Based.} 
In classical SLAM or structure from motion scenarios the camera is tracked in an unknown environment and a corresponding sparse 3D map or reconstruction of the environment is built. Each 3D point in the map has a corresponding image feature descriptor associated to it. Therefore, the main component of these methods is the detection of key-points in a query image and feature extraction, e.g. SIFT features, at these respective points. 2D to 3D point correspondences between the image and the 3D model can then be established using feature descriptor matching. 
%Given a 3D model and corresponding features associated with points of the model, the key component of these methods is the detection of key-points in a query image and feature extraction, e.g. SIFT features, at these points. 2D to 3D point correspondences between the image and the 3D model can then be established using feature matching. 
Finally, given these correspondences, the camera pose can be computed by solving the perspective-n-point problem. However, despite usually providing good camera localization accuracy, these methods can easily fail in case of texture-less surfaces and require efficient feature matching techniques to achieve reasonable computational times for camera re-localization applications. 
For this purpose, Sattler et al. \cite{sattler2011fast,sattler2012improving,sattler2017efficient} propose an optimized prioritization scheme based on vocabulary-based quantization for efficient feature matching. Additionally, by using co-visibility constraints or semantic consistency checks \cite{toft2018semantic}, wrong matches can be removed, which further improves the methods accuracy.
In contrast, Schmidt et al. \cite{schmidt2017self} focus on optimizing extracted features used for correspondence matching. Here, a deep learning method is applied and a neural network is trained on a contrastive loss function, pushing features of pixels to be similar only if they correspond to the same 3D point.
Implicitly giving a mapping between image pixels and 3D points, Shotton et al. \cite{shotton2013scene} train a regression forest on RGB and depth features extracted at pixel locations to estimate the corresponding scene coordinate directly. Further extensions and analysis of this method have been proposed, including uncertainty of the forests predictions \cite{valentin2015exploiting}, online adaption of the regression forest \cite{cavallari2017fly}, ensemble prediction \cite{guzman2014multi}, backtracking schemes \cite{meng2017backtracking} and a comparison to neural networks \cite{massiceti2017random}.
Switching from regression forests to convolutional neural networks, Brachmann et al. \cite{brachmann2017dsac,brachmann2018learning} propose an end-to-end trainable pipeline, consisting of a scene coordinate regression and a pose hypothesis scoring CNN, connected by a differentiable version of RANSAC, which they call DSAC. 
These methods have shown remarkable results in retrieving accurate camera poses. They, however, usually require depth information or a 3D model.

\paragraph{Image-Retrieval-Based.} In contrast to correspondence-based methods, image retrieval methods focus on computing a lower dimensional representation of a full query image, which can then efficiently be matched against a database of images with corresponding camera poses. Finding the nearest neighbor according to the resulting features will in this case also retrieve the closest camera pose, but therefore also restricts the search space to the camera poses contained in the database, especially if RGB images are the only source of information available. 
Glocker et al. \cite{glocker2015real} rely on a fern-based encoding approach, which computes a binary encoding for each frame and thus enables fast similarity comparisons based on the Hamming distance.
Relja et al. \cite{arandjelovic2016netvlad} construct a new feature aggregation layer, inspired by VLAD \cite{jegou2010aggregating}, which can be included in any existing convolutional neural network and shows great capabilities in aiding image retrieval tasks in the context of camera pose estimation.
Additionally, Taira et al. \cite{taira2018inloc} propose to learn dense features using a convolutional neural network for camera pose estimation. After retrieval of nearest neighbor database images, dense features at different layers of the network are used to find 2D-3D correspondences, given that depth information is available for the database images, from which the pose can be computed. Additionally, the estimated pose is verified by comparing the query image to the synthesized view obtained using the 3D model and retrieved camera pose.
\paragraph{Direct Pose Regression.} Very recently, direct camera pose regression approaches have emerged, mainly using CNNs to estimate the rotation, most often represented as quaternions, and position of the camera given a single RGB image as input. 
Starting with the introduction of PoseNet \cite{kendall2015posenet}, Kendall et al. presented a computationally very fast solution for solving the camera pose estimation problem relying solely on RGB information and also showing great capabilities when applied on large-scale scenes. This, on the other hand, came at a large drop in general accuracy compared to earlier state-of-the-art methods.
Thus, several extensions and modifications of this method have been proposed, including uncertainty estimation \cite{kendall2016modelling,kendall2017geometric,caihybrid}, LSTM units \cite{walch2017image}, frame-to-frame information \cite{balntas2018relocnet,laskar2017camera,clark2017vidloc,brahmbhatt2018geometry} and previous pose fusion \cite{valada2018deep,radwan2018vlocnet++}. The latter, however, more closely resembles a camera tracking scenario rather than re-localization as the method relies on pose information of the previous frame.

%\todo{Shouldn't we talk about Pose Refinement?}
%\todo{plz read the next paragraph carefully and try to make it consistent with our claims and ontributions. I didn't like the "side effect"}
Since in this work, we want to explore re-localization methods utilizing RGB information only, we build on top of recent research on direct camera pose regression methods. However, we additionally attempt to model the connection between an RGB image and its camera pose implicitly, rather than trying to simply learn this mapping directly. For this purpose, we show the advantage that leveraging an adversarial network can have on such methods. To the best of our knowledge, we are the first to investigate adversarial learning in the context of camera pose estimation. Therefore, we propose a novel framework based on a camera pose regression network and a discriminator network that, given a regressed pose and the RGB input image, learns to distinguish between regressed and ground truth poses. Further, once the model has learned a representation of this connection, as our main contribution, we show how the trained model can be used for camera pose refinement. By leveraging the learned information encoded in the discriminator network, the localization accuracy can be improved beyond the one of a simple camera pose regression network. 
\section{Methodology}
Following previous camera pose regression approaches, we attempt to train a convolutional neural network, hereby referred to as the pose regressor, to learn the mapping $\Omega:\textbf{x} \rightarrow \textbf{p}$ between an input image $\textbf{x}$ and a camera pose $\textbf{p}$.
%However, directly regressing the camera pose can quickly lead to the model "memorizing" the training data instead of learning a reasonable connection between input image and corresponding camera pose and as a result can lead to poor generalization capabilities of such models.

However, we additionally attempt to learn the distribution of camera poses and their respective RGB images captured by the camera. More precisely, we train a pose discriminator network to distinguish between regressed and ground truth pose with respect to the input image. The pose regressor and discriminator are trained in an alternating manner, where the pose regressors goal is to fool the discriminator, such that it can not clearly distinguish between regressed and real camera poses anymore. Finally, once the discriminator has learned the geometric mapping between an input image and a camera pose, the information captured by the discriminator can be leveraged to update and refine the regressed camera pose. By freezing the discriminator networks weights and optimizing solely the regressed camera pose, we aim at pushing the regressed pose closer towards the manifold of real poses to ultimately better fit the input image. An overview of our method can be seen in Figure \ref{fig:framework}.

\subsection{Camera Pose Regression}
Given an RGB image $\textbf{x} \in \mathbb{R^{\text{h} \times\text{w} \times \text{3}}}$, our objective is to predict the camera pose $\textbf{p} = [\textbf{q}, \textbf{t}]$ given as orientation, represented as vector $\textbf{q}$, and translation $\textbf{t} \in \mathbb{R^\text{3}}$. For this aim, a CNN, is trained on the following loss function
%\begin{equation}
%	\mathcal{L}_{pose} = \|\textbf{t}-\hat{\textbf{t}}\| + \beta \|\textbf{q}-\frac{\hat{\textbf{q}}}{\|\hat{\textbf{q}}\|}\|,
%	\label{Eq:PoseLoss}
%\end{equation}
\begin{equation}
\mathcal{L}_{pose} = \|\textbf{t}-\hat{\textbf{t}}\|e^{-\beta} + \beta + \|\textbf{q}-\hat{\textbf{q}}\|e^{-\alpha}+\alpha,
\label{Eq:PoseLoss}
\end{equation}

\noindent where $\hat{\textbf{t}}$ and $\hat{\textbf{q}}$ represent the predicted translation and rotation, respectively, $\beta$ and $\alpha$ are trainable parameters to balance both distances, and $\| \cdot \|$ is chosen to be the $\ell_1$ norm. Readers are referred to \cite{kendall2017geometric} for further details about the loss function, and its derivation. 

The parameterization used to regress the rotational component of an object or a camera pose has been extensively addressed in many literature \cite{brahmbhatt2018geometry,lienetbmvc2018}. In this work, first, we choose to evaluate our method on the representation of quaternions, which is already well established in image-based localization. Here, a quaternion can be described as $\textbf{q} = [w,\textbf{u}] \in \mathbb{R^\text{4}}$ where $w$ is a real valued scalar and $\textbf{u} \in \mathbb{R^\text{3}}$. To ensure that the resulting quaternions lie on the unit sphere, they are normalized during the training. As shown in \cite{kendall2017geometric}, no additional constraints are enforced while training the pose regression network, as the resulting quaternions become sufficiently close to the ground truth so that there is no significant difference in $\ell_1$ norm and spherical distance.
Second, we use the logarithm of a unit quaternion, which is computed as
\begin{equation}
	\textbf{q}_{log} =\text{log}~\textbf{q} = \begin{cases}
	\frac{\textbf{u}}{\|\textbf{u}\|} \arccos(w),& \text{if } \|\textbf{u}\| \neq 0\\
	\textbf{0},              & \text{otherwise}
	\end{cases}, 
\end{equation}
and has the advantages of not being over-parameterized. Further, it relaxes the need of normalization during the training. The the unit quaternion can be recovered by $\textbf{q}=[\cos(\|\textbf{q}_{log}\|), \frac{\textbf{q}_{log}}{\|\textbf{q}_{log}\|}sin(\|\textbf{q}_{log}\|)]$.
\subsection{Discriminator}
Both the regressed poses $\hat{\textbf{p}}$, and the ground-truth poses $\textbf{p}$, and a lower dimensional representation, $f(\textbf{x})$, of the corresponding input images, form "fake" and "real" examples, respectively, used to train the discriminator network. The aim of this network is to minimize the following loss function defined as
\begin{equation}
	\mathcal{L}_{D} = \sigma(\{f(\textbf{x}), \textbf{p}\}, c_{real}) + \sigma(\{f(\textbf{x}), \hat{\textbf{p}}\}, c_{fake}) ,
\end{equation}
where $\sigma(\cdot,\cdot)$ is the binary cross-entropy loss, $c_{real}$ and $c_{fake}$ are set to $1$ and $0$, respectively. Therefore, the discriminator models the conditional distribution $P(y|\ \textbf{p}, f(\textbf{x}))$ of $y \in \{c_{real}, c_{fake}\}$ conditioned on the pose $\textbf{p}$ and image features $f(\textbf{x})$, and thus implicitly captures the joint distribution of $\textbf{p}$ and $\textbf{x}$. Our framework is, in fact, inspired by GANs to ensure that the geometric mapping between camera poses and the corresponding RGB images are exploited in the network, however differs from the original GAN framework as our pose regression network is purely discriminative.
%It is important to note that training a discriminator to solely distinguish between real and fake poses, without providing information about the input image, is not beneficial, see Section \ref{sec:influencefeatures}, as it will simply model the training distribution, i.e. it will only memorize the trajectory of camera poses in the training set. Thus, it is unlikely to generalize to camera poses in the test set, which are not necessarily close to the camera pose distribution of the training dataset. 

\subsection{Feature Extraction}
A pre-trained network architecture on ImageNet \cite{ILSVRC15}, see Section \ref{sec:influencefeatures}, is used to extract a feature representation $f(\textbf{x})$ given an RGB input image. The weights of the network are frozen during the training, as its purpose is mainly to provide the discriminator with a lower dimensional representation of the image. Given the fact that most of the state-of-the-art network architectures produce a rather high dimensional feature representation (compared to the six or seven dimensional camera pose vector), and inspired by the concept of dimensionality reduction, we apply a linear mapping to better balance the dimensionality between feature representation and camera pose. 
%that are currently most often used, the resulting feature vector would still be overrepresented in comparison to the merely six or seven dimensional camera pose vector. Thus, similar to the concept of dimensionality reduction techniques, we apply a linear mapping to better balance the dimensionality between feature representation and camera pose. 
To easily integrate this linear mapping to the network architecture, we simply add one additional fully-connected layer, without bias or activation function, right after the last layer, and keep its weights frozen during training. This way, the discriminator is discouraged to adapt the extracted features during training and solely base its decision on the features themselves. 
% further reduce the dimensionality of the resulting feature, we add one additional fully-connected layer, without bias, to the extracted features, which is kept fixed during training as well.  This layer simply serves as a linear mapping to better balance the dimensionality between feature representation and camera pose. %This manual feature extraction is vital% as we always feed in the real images paired with either real or fake poses 
%, such that the discriminator is actually discouraged to learn meaningful image features to base its decision on and thus would not leverage any geometric information.
%The extracted features are then concatenated with the camera pose and used as input for the discriminator network.
The camera pose vector is then copied, to fit the dimensionality of the extracted feature representation, and concatenated with said representation to form a feature map that is used as the input to the discriminator network.
Intuitively we would want the discriminator to learn the connection between RGB images and corresponding camera poses. Therefore, such that the network is discouraged to solely focus on the information provided by either one, the design choices described above were made. However, %to balance the given information and therefore steer the learning process, 
in addition we have experimented with fine-tuning the feature extraction network as well as only fine-tuning individual layers. Both resulted in worse performance. 
\subsection{Adversarial Learning}
Following the training procedure introduced for generative adversarial networks, we alternate between training the camera pose regressor and the discriminator network, updating the regressor on
\begin{equation}
\label{eq:L_G}
	\mathcal{L}_{G} = \mathcal{L}_{pose} + \lambda \underbrace{\sigma(\{f(\textbf{x}), \hat{\textbf{p}}\}, c_{real})}_{\mathcal{L}_{adv}},
\end{equation}
such that the network learns to predict more and more realistic poses and thus eventually is able to fool the discriminator. Here, the parameter $\lambda$ balances the influence of the adversarial loss on the pose regressor. %\todo{don't we need to birdge the gap between the $L_D$ and $L_G$}
\subsection{Pose Refinement} 
\label{sec:advref}
Once the model is trained and the discriminator is successfully "fooled", meaning it can not distinguish properly between regressed and ground truth poses with respect to the input image, the discriminator network can be used during testing to refine the regressed camera poses. For this aim, the test image is fed to the pose regression network to obtain an initial pose estimate. Then, the predicted pose together with the extracted feature representation of the image is used as input to the discriminator. In succession, however, the weights of the discriminator are frozen, and the initially regressed pose $\hat{\textbf{p}}$ for the image $\textbf{x}$ is updated  iteratively by minimizing the loss function as
\begin{equation}
	\mathcal{L}_{ref} = \sigma(\{f(\textbf{x}), \hat{\textbf{p}}\}, c),
\end{equation}
where the class label $c$ is set to $0.5$. % as we aim to update the pose, such that the discriminator is not able to correctly classify this sample. 
This stems from the fact, that at the end of training, the discriminator will not be able to distinguish between regressed and ground truth camera pose anymore, thus predicting values close to $0.5$ in both cases. Intuitively, this amounts to moving along the manifold towards a region where the discriminator reliably confuses real and regresses poses. Therefore, any predicted pose of an unseen query image should be pushed towards this manifold.
As the gradients coming from the discriminator do not necessarily follow a geometrically meaningful direction, in case of using the quaternion representation, we restrict the quaternion update, so that its movement along the unit sphere is ensured \cite{byrne2013geodesic, birdal2018bayesian}. Thus, the update for one iteration is described by

\begin{equation}
	\textbf{q}_t = \textbf{q}_{t-1} \cos(\gamma l) + \frac{\textbf{v}}{\gamma} \sin(\gamma l) ,
\end{equation}

\noindent with $\gamma = \| \textbf{v} \|_2$, $l$ being the step size, and $\textbf{v} \in \mathbb{R^\text{4}}$ being the projection of the quaternion gradient $\nabla \textbf{q}$ into the tangent space, given as

\begin{equation}
	\textbf{v} = (I -\nabla \textbf{q} \nabla \textbf{q}^T)\nabla \textbf{q},
\end{equation}
where $I \in \mathbb{R^{\text{4} \times \text{4}}}$ is the identity matrix. To further ensure that the resulting poses are valid, the updated quaternion is normalized after each iteration. However, no such constraints have to be be enforced to update the translational component of the camera pose. Though, for simplicity, it is updated with the same step size $l$.

\subsection{Training Procedure}
\label{sec:training}
As a first step, the pose regression network is trained for a few epochs to initially give reasonable poses, before including the adversarial loss in the training procedure, where the parameters $\beta$ and $\alpha$ are set following the state-of-the-art \cite{brahmbhatt2018geometry} and $\lambda$ is set to $1 \cdot 10^{-3}$. Afterwards, the pose regressor and discriminator are alternately trained on the $\mathcal{L}_{G}$ and $\mathcal{L}_{D}$ loss functions, respectively.
%Instead of injecting random noise into the pose regressor, as originally required for the conditional GAN framework, we placed dropout layers at a dropout rate $0.2$ before every convolutional layer and keep them active both during training and testing. This strategy has already been employed for Image-to-Image translation \cite{pix2pix2017}, where it has been pointed out that explicit noise would simply be ignored by the model.\\ 
% Is that true? not anymore
\paragraph{Implementation Details.}
Following the state-of-the art \cite{brahmbhatt2018geometry}, input RGB images are down-sampled to a resolution of 341 $\times$ 256 pixels, normalized, and then fed in mini batches of size 64 to train the neural networks. As a camera pose regressor, a ResNet-34 network architecture is used as the base network, where the classification layers are removed and two fully connected layers for camera pose regression are placed after the average pooling layer. %During training, after experimental evaluation, the parameter $\beta$ of the pose loss $\mathcal{L}_{pose}$ from Eq.~\ref{Eq:PoseLoss} is set to $1$. We refer the reader to \cite{kendall2015posenet} and \cite{kendall2017geometric} on how to suitably choose this parameter, as especially for larger scenes the quaternion and translation losses need to be properly balanced due to their very different scale. As the scenes of the dataset we evaluate on are still rather small, we did not find a significant change in accuracy and thus for simplicity chose this parameter value. It could, however, be tuned for each scene, which we would expect to improve the scene specific performance a little.
The discriminator consists of three convolutional layers followed by exponential linear units as activation function. All networks are implemented in PyTorch. For training the networks, we use the Adam Optimizer with a learning rate of $1 \cdot 10^{-4}$ and optimize for 300 epochs on an 11GB NVIDIA GeForce RTX 2080 graphics card. Once the networks are trained, the regressed camera poses are refined as described in Section \ref{sec:advref} until convergence, but up to a maximum of 50 iterations at a step size of $l = 1 \cdot 10^{-3}$. The effect of the step size and the number of iterations on the resulting pose accuracy can also be found in more detail in Section \ref{sec:advrefres}.
\section{Experiments and Evaluation}
\begin{table}[t]
	\begin{center}
		\caption{Effect of adversarial training and pose refinement on the camera pose accuracy, evaluated on the \textit{Heads} scene. Median rotation and translation errors are reported. Optimizing the camera pose regression network with the adversarial loss results in an improvement in accuracy, which is further increased by our proposed camera pose refinement.}
		\resizebox{0.49\textwidth}{!}{
			\begin{tabular}{lccc}
				Scene & Base Model& Ours & Ours+Ref.\\
				\noalign{\smallskip}
				\midrule
				\noalign{\smallskip}
				%\multirow{2}{*}{Heads}  & $15.1^\circ$ & $14.4^\circ$ & $11.2^\circ$ \\
				%& \SI{0.24}\meter & \SI{0.23}\meter & \SI{0.22}\meter \\
				\textit{Heads}  & $14.5^\circ$, \SI{0.18}\meter & $14.1^\circ$, \SI{0.17}\meter & $12.4^\circ$, \SI{0.16}\meter \\
				%& \SI{0.18}\meter & \SI{0.17}\meter & \SI{0.16}\meter \\
				%\midrule
				%\multirow{2}{*}{7-scenes} &  &  & \\
				%&  &  &  \\
				\bottomrule
			\end{tabular}
		}
		%\vspace{0.3cm}
		\label{table:adv}
	\end{center}
\end{table}
\begin{figure*}[t]
	\centering
	\begin{subfigure}[b]{0.44\textwidth}
		\includegraphics[width=\textwidth]{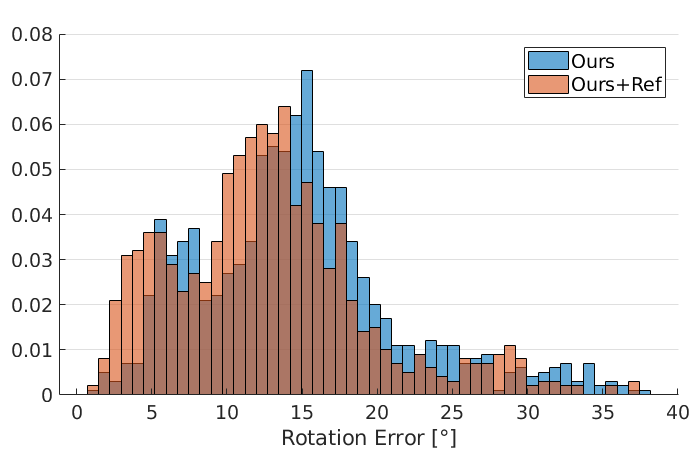}
	\end{subfigure}
	\hspace{1.0cm}
	\begin{subfigure}[b]{0.44\textwidth}
		\includegraphics[width=\textwidth]{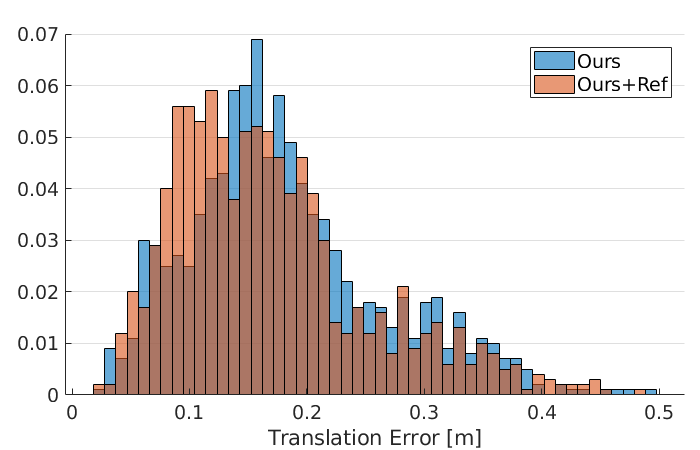}
	\end{subfigure}
	%\begin{subfigure}[b]{0.24\textwidth}
	%	\includegraphics[width=\textwidth]{figures/stairs_rotation.png}
	%\end{subfigure}
	%\hspace{1cm}
	%\begin{subfigure}[b]{0.24\textwidth}
	%	\includegraphics[width=\textwidth]{figures/stairs_translation.png}
	%\end{subfigure}
	%\begin{subfigure}[b]{0.24\textwidth}
	%	\includegraphics[width=\textwidth]{figures/fire_translation.png}
	%\end{subfigure}
	\caption{Normalized histograms of rotation and translation errors before and after pose refinement on the \textit{Heads} scene. Results without refinement (Ours) are shown in blue, whereas errors after refinement (Ours+Ref.) are displayed in orange, resulting in an overlap in brown.}
	\label{fig:hists}
\end{figure*}
We evaluate our method on the publicly available 7-Scenes \cite{shotton2013scene} dataset.
This dataset from Microsoft consists of RGB-D frames of seven indoor scenes, captured with a hand-held Kinect camera, and corresponding ground truth camera poses computed using Kinect Fusion. The scenes are of varying spatial extent and also differ significantly in the amount of training data available. Training and test data are specified and consist of distinct camera trajectories. It has been widely used to evaluate camera re-localization methods as it contains several challenging scenarios such as motion blur, repeating structures and texture-less surfaces.\\
%\textbf{Cambridge Landmarks.} Containing six outdoor scenes located around Cambridge Univerity, this dataset provides RGB images for training as well as test data, captured using a mobile phone and ground truth camera pose computed using structure from motion. 
 
For evaluation, we utilize the recent state-of-the-art method and implementation of MapNet \cite{brahmbhatt2018geometry}, focusing on directly regressing the camera pose without the aid of temporal or geometric information. We investigate the effect of our method on models either regressing quaternions themselves or the logarithm of a quaternion (baseline models of \cite{brahmbhatt2018geometry}). 
Further, for evaluation of our framework, we introduce the following models:
\begin{itemize}
\item{\textbf{Baseline:}}
As a baseline model, we train the camera pose regression network on the $\mathcal{L}_{pose}$ loss, which, as already mentioned, effectively results in the state-of-the art baseline method of \cite{brahmbhatt2018geometry}. However, we abbreviate this model as ``Base Model'' whenever experiments are conducted by us to explicitly highlight re-trained models and to better analyze the effect of our contributions.
\item{\textbf{Adversarial Pose Regression:}}
To analyze the effect of adversarial training on the camera pose regression, the regression model is trained on the $\mathcal{L}_{G}$ loss function (Eq.\ref{eq:L_G}),
%, which includes a pose regression losee $\mathcal{L}_{pose}$ as well as an adversarial loss term $\mathcal{L}_{adv}$, 
abbreviated as ``Ours''.
\item{\textbf{Pose Refinement:}}
Finally, during testing, the trained discriminator network is used to further improve the regressed poses using $\mathcal{L}_{ref}$. The models are then abbreviated as ``Ours+Ref''. 
% the results of which are summarized by the models abbreviated as ``Ours+Ref''.
\end{itemize}
In the remainder of this section, these models will be used to validate our contributions. We start by investigating the effect of optimizing a camera pose regression network including the adversarial loss, after which we analyze the effect of the proposed pose refinement on the localization accuracy. Finally, setting our method in the context of recent research, we compare our results to the current state-of-the-art methods on direct camera pose regression.
\begin{figure*}[t]
	\begin{center}
		\begin{subfigure}[b]{0.21\textwidth}
			\includegraphics[width=\textwidth]{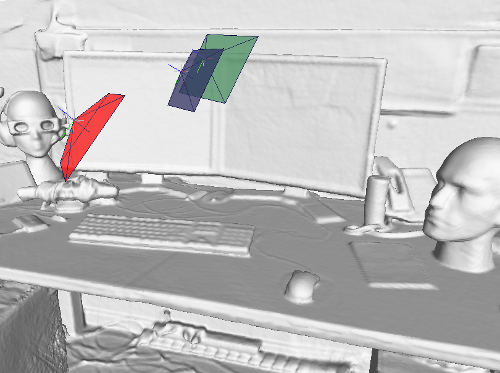}
			\subcaption{$15.3^\circ$-$11.0^\circ$, $9$cm-$6$cm}
			\vspace{0.1cm}
		\end{subfigure}
		\begin{subfigure}[b]{0.21\textwidth}
			\includegraphics[width=\textwidth]{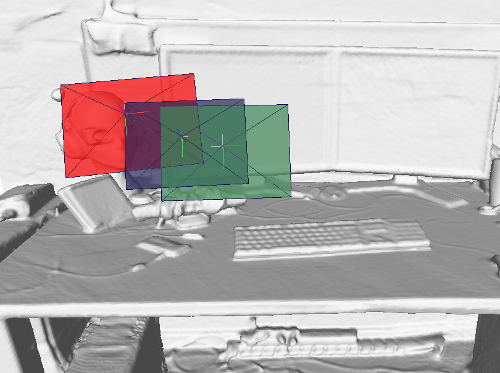}
			\subcaption{$8.8^\circ$-$3.1^\circ$, $9$cm -$6$cm}
			\vspace{0.1cm}
		\end{subfigure}
		\begin{subfigure}[b]{0.21\textwidth}
			\includegraphics[width=\textwidth]{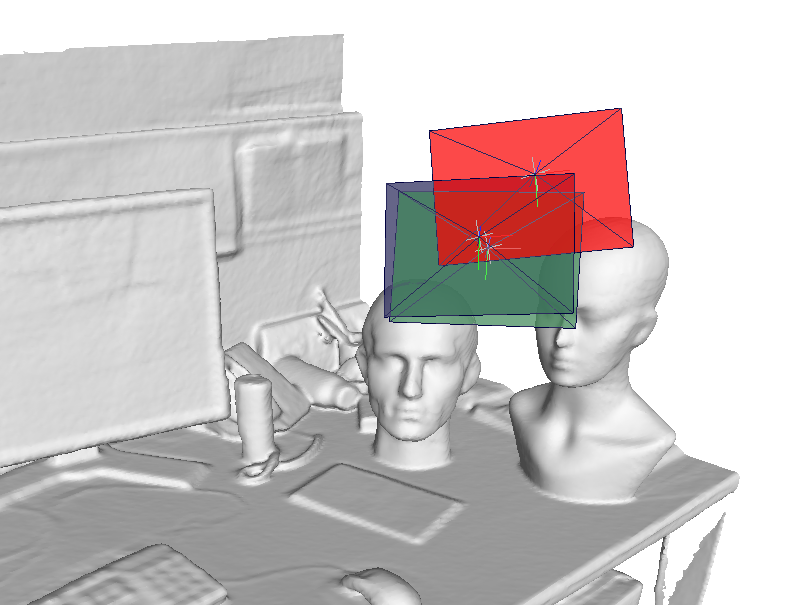}
			\subcaption{$8.0^\circ$-$3.6^\circ$, $5$cm-$1$cm}
			\vspace{0.1cm}
		\end{subfigure}
		\begin{subfigure}[b]{0.21\textwidth}
			\includegraphics[width=\textwidth]{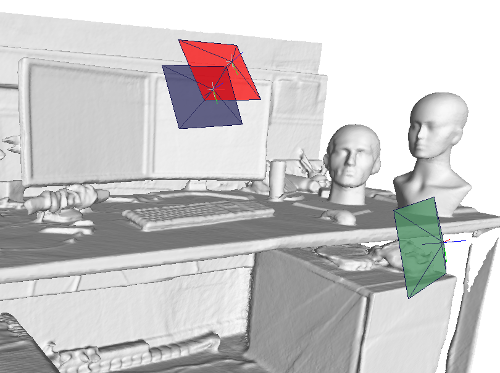}
			\subcaption{$15.7^\circ$-$11.5^\circ$, $36$cm-$37$cm}
			\vspace{0.1cm}
		\end{subfigure}
		\hfill	
		\includegraphics[width=0.21\textwidth]{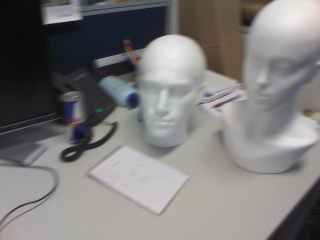}
		\includegraphics[width=0.21\textwidth]{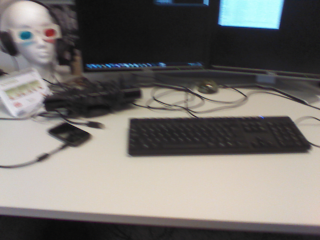}
		\includegraphics[width=0.21\textwidth]{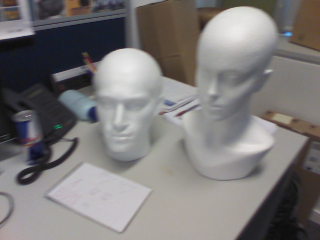}
		\includegraphics[width=0.21\textwidth]{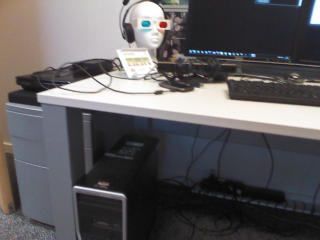}
	\end{center}
	\caption{RGB input images (second row) and the corresponding camera poses (first row), visualized in a reconstruction of the given scene. For each frame, the ground truth (\textcolor{green}{green}), initially regressed pose (\textcolor{red}{red}) and optimized pose using the proposed refinement (\textcolor{blue}{blue}) are displayed. Below each visualization the respective rotation and translation errors before and after refinement are given.}
	\label{fig:vis2}
\end{figure*}
\begin{figure}[t]
	\begin{center}
		\begin{subfigure}[b]{0.47\textwidth}
			\includegraphics[width=\textwidth]{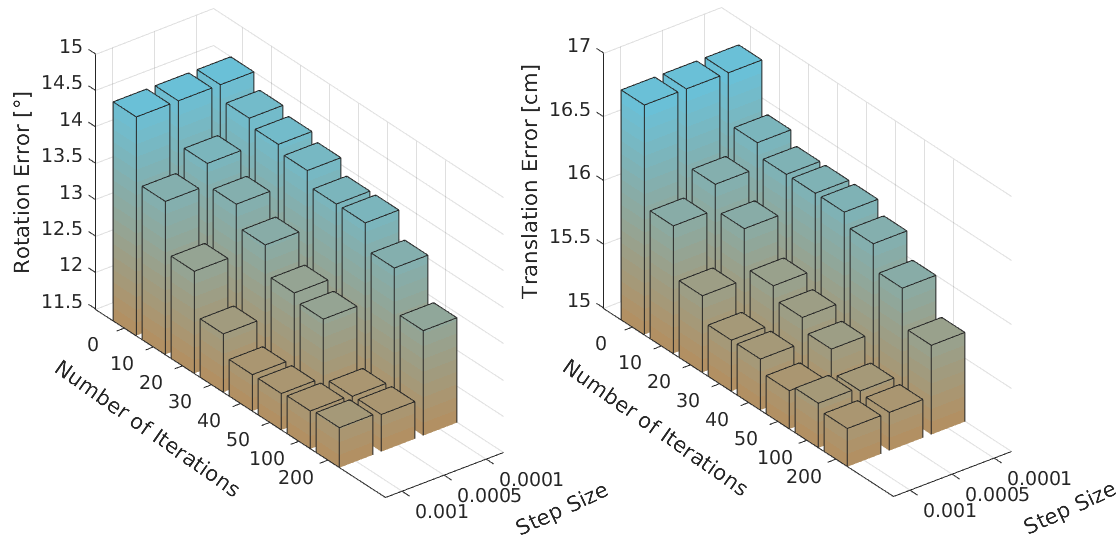}
			%\caption{Heads}
		\end{subfigure}
		%~
		%\begin{subfigure}[b]{0.49\textwidth}
		%	\includegraphics[width=\textwidth]{figures/stairs_refinement.png}
		%\caption{Stairs}
		%\end{subfigure}	
	\end{center}
	\caption{Effect of different numbers of iterations as well as step sizes on the median rotation and translation errors for the proposed refinement, shown on the \textit{Heads} scene. Our refinement can significantly improve the localization accuracy even in a few iterations of optimization.}
	\label{fig:refinement}
	\vspace{-0.1cm}
\end{figure}

\subsection{Ablation Studies}
\label{sec:ablationstudy}
\subsubsection{Adversarial Learning}
First, to investigate the effect of adversarial learning on the camera pose regression framework, we compare rotation and translation errors of our baseline, ``Base Model'', and the model ``Ours''. The results can be seen in Table \ref{table:adv}, showing median rotation and translation errors of the described models on the \textit{Heads} scene. %\todo{why we are bringing the next paragraph here?} 
That adversarial training can help in training deep networks has already been shown, for example in \cite{yang20183d} for the task of human pose estimation, which, however differs significantly from the task of predicting the camera pose from a corresponding image. Nevertheless, we found slight improvements in rotation, as well as in translation accuracy by simply including adversarial training into a camera pose regression framework due to better and more stable convergence of the model. 
%We want to note that this effect depends on the choice of $\lambda$ and if not tuned can actually harm the resulting localization accuracy as well due to increased overfitting to the training set. We therefore experimentally fine-tuned $\lambda$ to find an optimal balance between pose and adversarial loss and set it to $1 \cdot 10^{-3}$, as described in Section \ref{sec:training}. In addition more information on setting this parameter can be found in the supplementary material.
\subsubsection{Pose Refinement}
\label{sec:advrefres}
As a second step, we evaluate our proposed pose refinement based on the trained discriminator network. Surprisingly, even though the gradients coming from the discriminator have not specifically been trained to have geometric meaningful information, it turns out that this information has implicitly been encoded in the network. Thus, we can use the gradients to update the regressed poses for any test image, given the constraints described in Section \ref{sec:advref} on the quaternion update. Table \ref{table:adv} and Figure \ref{fig:hists} summarize our findings, where we report the median rotation and translation error as well as the overall distribution of the aforementioned errors on the \textit{Heads} scene of the 7-scenes dataset. Overall we found improvements in pose accuracy by applying the proposed pose refinement, examples of which are also visualized in Figures \ref{fig:vis} and \ref{fig:vis2}. Further examples for the remaining scenes of the 7-Scenes dataset can be found in the supplementary material. It can be seen both quantitatively and qualitatively that the regressed pose can effectively be pushed further towards the ground truth pose by the proposed refinement step, resulting for example in a relative improvement in rotation of $12.0\%$ and $31.1\%$ for the Heads and Stairs scene respectively.  
%In comparison to our baseline model Ours$_{base}$, we can report a relative improvement of $9.3\%$ in translation and $2.9\%$ in rotation. Although, on average this improvement might seem rather small, it should be noted that the effect of our method seems to be most noticeable on scenes for which a much smaller number of training images is available, such as \textit{Heads}, \textit{Stairs} and \textit{Fire} with only one and two thousand training images available in comparison to the largest training set of seven thousand images.

Further, we investigate the effect of the step size $l$ as well as the number of iterations on the localization accuracy of the proposed pose refinement. The results of our investigation are summarized in Figure \ref{fig:refinement}, where we show median rotation and translational error on the \textit{Heads} scenes for different numbers of refinement iterations as well as step sizes. A lower step size usually leads to smaller changes in the pose, but, therefore, can also require a higher number of iterations to converge to the desired pose. Since this optimization process is required during testing, increasing the number of iterations is directly proportional to an increase in computational time. Experiments with larger step sizes ($l >10^{-3}$) resulted in deterioration of the camera poses due to the optimization procedure becoming unstable.
%However, as can be seen in Figure \ref{fig:refinement}, a higher step size can possibly harm the final pose accuracy as well. 
Usually only a few iterations of refinement are sufficient, though, to improve the regressed poses and provide a good improvement in camera pose accuracy, whereas the run-time of RANSAC-based methods, for example, depends on the quality of correspondences found. As a trade-off, we chose the parameter setting described in Section \ref{sec:training}. For example, on average the refinement has a computational time of $42$ms for $30$ iterations, but grows linearly with the number of iterations. Although we were able to achieve promising results with the proposed pose refinement strategy, it should be noted that it remains an optimization procedure itself, and thus depends on factors such as the quality of initialization. Therefore, in some cases the refinement might result in a solution that is not preferable to the initially regressed pose or difficult to recover from, if the predicted pose is far away from the ground truth one, an example of which is shown in Figure \ref{fig:vis2} d). 

\begin{table*}[t]
	\begin{center}
		\caption{Comparison between recent state-of-the-art direct camera pose regression methods and our results without (Ours) and with pose refinement (Ours+Ref.). Following the state of the art, displayed is the median rotation and translation error evaluated on the 7-Scenes dataset.}
		\resizebox{\textwidth}{!}{		
			\begin{tabular}{l||ccccc||ccc}
				\toprule\noalign{\smallskip}
				Scene & \vtop{\hbox{\strut DSAC++}\hbox{\strut RGB \cite{brachmann2018learning}}}&\vtop{\hbox{\strut PoseNet}\hbox{\strut RGB \cite{kendall2017geometric}}} & \vtop{\hbox{\strut MapNet \cite{brahmbhatt2018geometry}}\hbox{\strut }} & Ours & Ours+Ref.& \vtop{\hbox{\strut MapNet \cite{brahmbhatt2018geometry}}\hbox{\strut ~~~~$\text{log q}$}} & \vtop{\hbox{\strut Ours}\hbox{\strut $\text{log q}$}} & \vtop{\hbox{\strut Ours+Ref.}\hbox{\strut ~~~~$\text{log q}$}}\\
				\noalign{\smallskip}
				\midrule
				\midrule
				\noalign{\smallskip}
				Chess & $0.02$m, $0.7^\circ$ &$0.14$m, $4.5^\circ$ &
				$0.11$m, $4.2^\circ$  & $0.13$m, $4.9^\circ$ & $0.12$m, $4.8^\circ$ & $0.11$m, $4.3^\circ$ & $0.13$m, $5.0^\circ$ &$0.12$m, $4.8^\circ$\\
				Fire & $0.03$m, $1.1^\circ$& $0.27$m, $11.8^\circ$ & $0.29$m, $11.7^\circ$ & $0.30$m,$11.0^\circ$& $0.29$m, $10.2^\circ$ & $0.27$m, $12.1^\circ$ & $0.28$m, $11.8^\circ$& $0.27$m, $11.6^\circ$\\
				Heads &$0.12$m, $6.7^\circ$& $0.18$m, $12.1^\circ$ &
				$0.20$m, $13.1^\circ$  & $0.17$m, $14.5^\circ$ & $0.15$m, $12.0^\circ$ & $0.19$m, $12.2^\circ$ & $0.17$m, $14.1^\circ$& $0.16$m, $12.4^\circ$\\
				Office &$0.03$m, $0.8^\circ$& $0.20$m, $5.7^\circ$ & $0.19$m, $6.4^\circ$ & $0.22$m, $6.7^\circ$ & $0.21$m, $6.6^\circ$ & $0.19$m, $6.4^\circ$ &$0.20$m, $7.1^\circ$ &$0.19$m, $6.8^\circ$\\
				Pumpkin &$0.05$m, $1.1^\circ$& $0.25$m, $4.8^\circ$ & $0.23$m, $5.8^\circ$ & $0.23$m, $6.7^\circ$ & $0.22$m, $6.5^\circ$ & $0.22$m, $5.1^\circ$ &$0.22$m, $5.4^\circ$ & $0.21$m, $5.2^\circ$\\
				Red Kitchen &$0.05$m, $1.3^\circ$& $0.24$m, $5.5^\circ$ & $0.27$m, $5.8^\circ$ & $0.27$m, $5.9^\circ$ & $0.26$m, $5.8^\circ$ & $0.25$m, $5.3^\circ$ &$0.26$m, $6.2^\circ$ & $0.25$m, $6.0^\circ$ \\
				Stairs &$0.29$m, $5.1^\circ$& $0.37$m, $10.6^\circ$ & $0.31$m, $12.4^\circ$ & $0.32$m, $13.5^\circ$ & $0.30$m, $12.2^\circ$ & $0.30$m, $11.3^\circ$ & $0.29$m, $12.2^\circ$ & $0.28$m, $8.4^\circ$\\
				\midrule
				Average &$0.08$m, $2.4^\circ$& $0.24$m, $7.9^\circ$ & $0.23$m, $8.5^\circ$  & $0.23$m, $9.0^\circ$ & $0.22$m, $8.3^\circ$ & $0.22$m, $8.1^\circ$  & $0.22$m, $8.8^\circ$ & $0.21$m, $7.9^\circ$\\
				%\midrule
				%\midrule
				%Great Court & - & - & - & $7.00m$, $3.65^\circ$ & - \\
				%King's College & $1.66m$, $4.86^\circ$ & $1.74m$, $4.06^\circ$ & $7.00m$, $3.65^\circ$ & $0.99m$, $1.06^\circ$ & -\\
				%Old Hospital & $2.62m$, $4.9^\circ$ & $2.57m$, $5.14^\circ$ & $1.51m$, $4.29^\circ$ & $2.17m$, $2.94^\circ$ & -\\
				%Shop Facade & & & & &\\
				%Mary's Church & & & & &\\
				%Street & & & & &\\
				%\midrule
				%Average & & & & &\\
				\bottomrule
			\end{tabular}
		}
		\vspace{-0.1cm}
		\label{table:results}
	\end{center}
\end{table*}

\subsubsection{Influence of Feature Extractor}
\label{sec:influencefeatures}
To evaluate the effect of the feature extraction network on the discriminator and thus the camera pose refinement, we evaluated our method using several different network architectures, namely AlexNet \cite{krizhevsky2012imagenet}, VGG16 \cite{simonyan2014very} and ResNet-18 \cite{he2016deep}. Initialization is kept the same for all models and refinement is run for $30$ iterations. 
Additionally we experiment with feeding only the regressed camera poses to train the discriminator network. For this experiment we replace the convolutional layers of the discriminator network with fully connected layers of roughly equal number of trainable parameters as the convolutional variant of the discriminator. Since a separate training is required for each architecture, we report the relative decrease in rotation and translation error over the initially regressed pose quality of the respective model. The results are summarized in Table \ref{table:nets}. We found that our proposed refinement is fairly robust to the extracted features and were able to obtain improved pose accuracy regardless of the network architecture used, except when using pose information only, without additional information about the corresponding image representation. Nevertheless, we found an increase in localization performance depending on the choice of network architecture with the best performing model resulting in the ResNet-18 \cite{he2016deep} network architecture.

\begin{table}[t]
	\begin{center}
		\caption{Relative decrease, in percentage, of the median rotation and translation error after refinement in comparison to initially regressed poses. Evaluated are different network architectures used to obtain a feature representation of the RGB image input, showing the influence of the feature extractor on the proposed refinement. Higher values correspond to improved pose accuracy.}
		\resizebox{0.48\textwidth}{!}{
			\begin{tabular}{lcccc}
				\textit{Heads} & \vtop{\hbox{\strut Without}\hbox{\strut $f(x)$}}  & \vtop{\hbox{\strut AlexNet}\hbox{\strut \cite{krizhevsky2012imagenet}}}  & \vtop{\hbox{\strut VGG-16}\hbox{\strut \cite{simonyan2014very}}} & \vtop{\hbox{\strut ResNet-18}\hbox{\strut \cite{he2016deep}}}\\
				\noalign{\smallskip}
				\midrule
				\noalign{\smallskip}
				%\multirow{2}{*}{Heads}  & $15.1^\circ$ & $14.4^\circ$ & $11.2^\circ$ \\
				%& \SI{0.24}\meter & \SI{0.23}\meter & \SI{0.22}\meter \\
				Rotation & 4.25\% & 3.56\% & 8.32\% & 12.18\%\\
				Translation & -3.0\% & 2.88\% & 4.7\% & 4.39\% \\
				%\midrule
				%\multirow{2}{*}{7-scenes} &  &  & \\
				%&  &  &  \\
				\bottomrule
			\end{tabular}
		}
		\vspace{-0.2cm}
		\label{table:nets}
	\end{center}
\end{table}

\subsection{Comparison to the State of the Art}
As our main focus in this work is to investigate the effect of our proposed framework on direct camera pose regression methods using RGB information only, we show a comparison to recent methods working on this topic, namely PoseNet \cite{kendall2017geometric} and MapNet \cite{brahmbhatt2018geometry}, which also forms our baseline model. We choose PoseNet and MapNet versions solely relying on single image and RGB information, for which we show the results in Table \ref{table:results}.
We evaluate both models trained to predict quaternions as well as the logarithm of quaternions to show the effectiveness of our method regardless of the baseline representation used. In comparison to both \cite{kendall2017geometric} and \cite{brahmbhatt2018geometry}, we found overall improvements in pose accuracy using the proposed refinement, where the effect of our method seems to be most profound on scenes for which only a small number of training images is available, such as \textit{Heads} and \textit{Stairs}. In addition we include a recent scene coordinate regression method, DSAC++ \cite{brachmann2018learning}, that given an initial depth estimate, can be trained solely relying on RGB information. As can be seen, the regressed 3D information, and following pose refinement, greatly improve the accuracy of the predicted camera poses, which leads to the method outperforming direct camera pose regression methods and ours. This, however, comes at a significant drop in computational time. 
Lastly, although we focus on RGB only solutions in this paper, it should be mentioned that our core regression method could be easily extended to include further information, like relative pose information or geometric constraints as in \cite{brahmbhatt2018geometry}.

\section{Conclusion}
In conclusion, we have presented a novel approach for camera re-localization applications solely relying on RGB information. Building on top of direct camera pose regression methods, we use the regressed camera poses and features extracted from the input image to train a discriminator network that tries to distinguish between regressed and ground truth poses, and thus implicitly tries to learn the geometric connection between RGB image and the corresponding camera pose. We have analyzed each component of our framework to evaluate this assumption and were able to achieve promising results. Further, we proposed a novel RGB-based pose refinement, where we use the trained discriminator network to update and optimize the initially regressed poses, showing that the network can learn a meaningful representation of the camera poses and image space, and in turn can use this information to further improve localization accuracy.
\pagebreak
\clearpage
\appendix

%%%%%%%%% TITLE
%\title{Supplementary Material}
%\maketitle
%\thispagestyle{empty}
%%%%%%%%% BODY TEXT
\begin{figure}[h]
	\begin{minipage}[t]{\textwidth}
		%\begin{center}
		%	\Large{\textbf{Adversarial Networks for Camera Pose Regression and Refinement - Supplementary Material}}\\
		%\end{center}
		
		\section{Qualitative Results}
		\captionsetup[subfigure]{labelformat=empty}
		\begin{center}
			\includegraphics[width=0.22\textwidth]{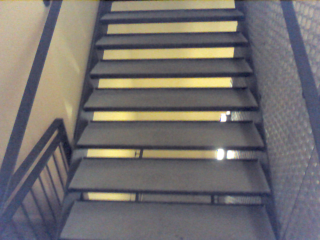}
			\includegraphics[width=0.22\textwidth]{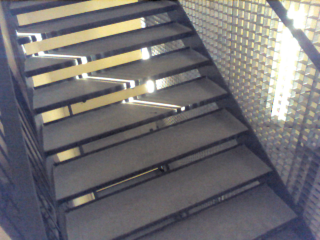}
			\includegraphics[width=0.22\textwidth]{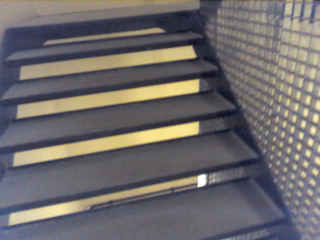}
			\includegraphics[width=0.22\textwidth]{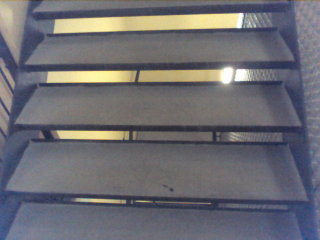}
			
			\begin{subfigure}[h]{0.22\textwidth}
				\includegraphics[width=\textwidth]{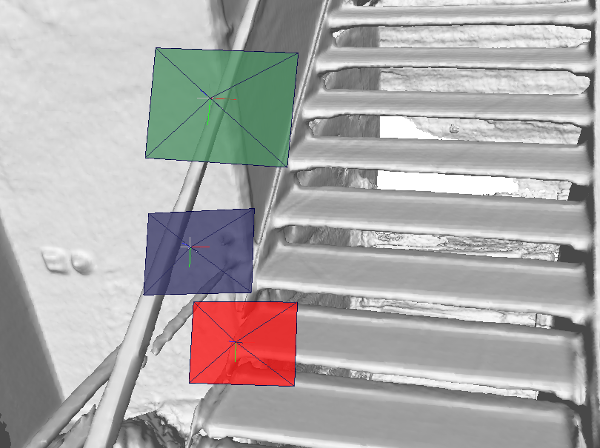}
				\subcaption{$6.9^\circ$-$1.5^\circ$, \SI{0.15}\meter -\SI{0.14}\meter}
			\end{subfigure}
			\begin{subfigure}[h]{0.22\textwidth}
				\includegraphics[width=\textwidth]{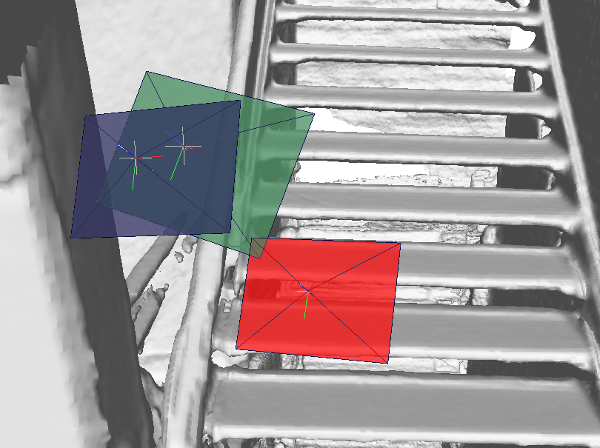}
				\subcaption{$13.9^\circ$-$9.2^\circ$, \SI{0.15}\meter -\SI{0.05}\meter}
			\end{subfigure}
			\begin{subfigure}[h]{0.22\textwidth}
				\includegraphics[width=\textwidth]{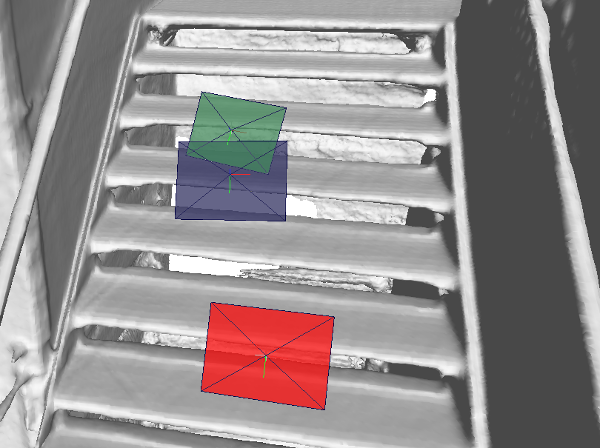}
				\subcaption{$13.4^\circ$-$9.5^\circ$, \SI{0.38}\meter -\SI{0.17}\meter}
			\end{subfigure}
			\begin{subfigure}[h]{0.22\textwidth}
				\includegraphics[width=\textwidth]{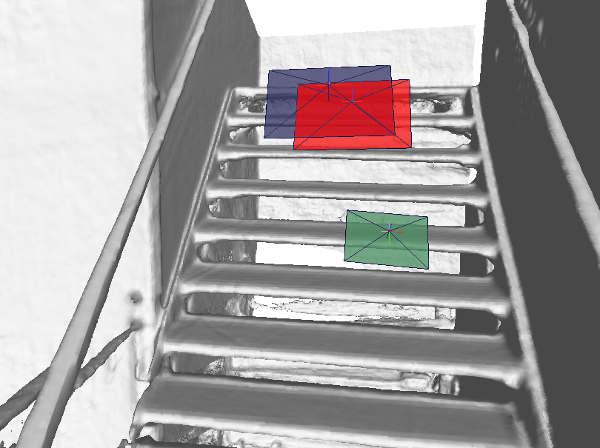}
				\subcaption{$7.4^\circ$-$3.0^\circ$, \SI{0.32}\meter -\SI{0.35}\meter}
			\end{subfigure}
			
			\vspace{0.2cm}
			\includegraphics[width=0.22\textwidth]{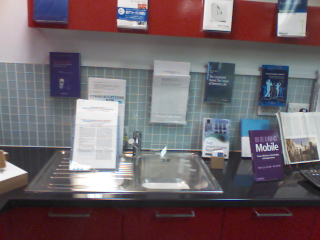}
			\includegraphics[width=0.22\textwidth]{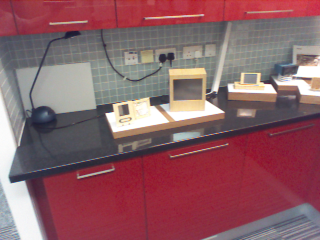}
			\includegraphics[width=0.22\textwidth]{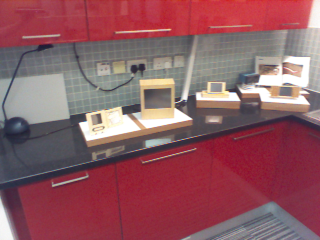}
			\includegraphics[width=0.22\textwidth]{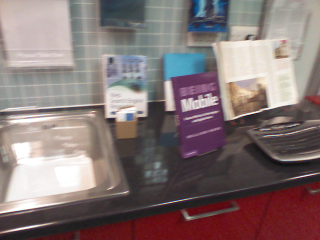}
			\begin{subfigure}[h]{0.22\textwidth}
				\includegraphics[width=\textwidth]{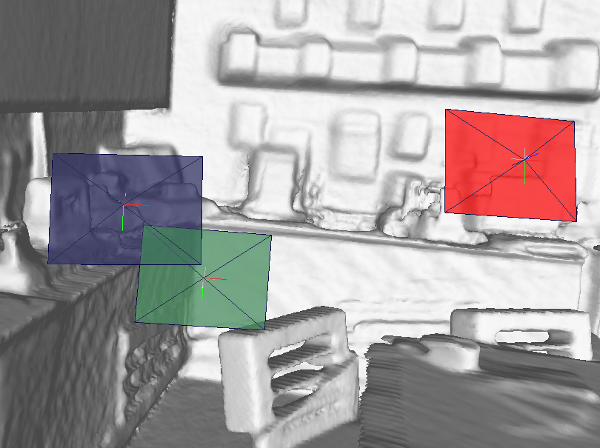}
				\subcaption{$2.9^\circ$-$2.2^\circ$, \SI{0.14}\meter -\SI{0.11}\meter}
			\end{subfigure}
			\begin{subfigure}[h]{0.22\textwidth}
				\includegraphics[width=\textwidth]{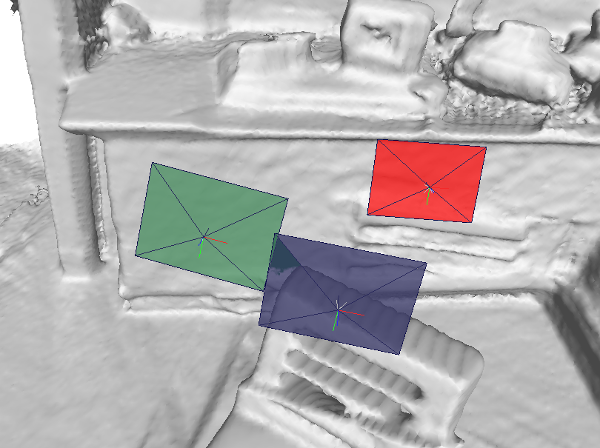}
				\subcaption{$15.4^\circ$-$3.9^\circ$, \SI{0.1}\meter -\SI{0.06}\meter}		
			\end{subfigure}
			\begin{subfigure}[h]{0.22\textwidth}
				\includegraphics[width=\textwidth]{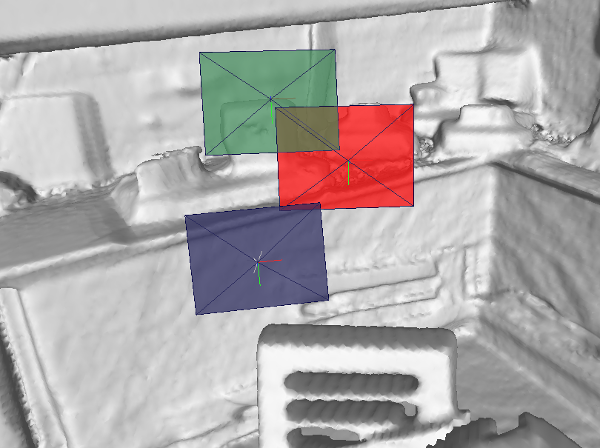}
				\subcaption{$6.7^\circ$-$2.3^\circ$, \SI{0.16}\meter -\SI{0.06}\meter}
			\end{subfigure}
			\begin{subfigure}[h]{0.22\textwidth}
				\includegraphics[width=\textwidth]{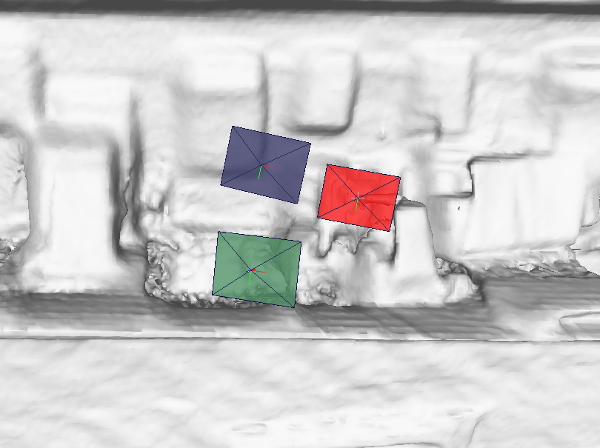}
				\subcaption{$5.0^\circ$-$3.7^\circ$, \SI{0.24}\meter -\SI{0.26}\meter}
			\end{subfigure}
			\hfill	
			
			\vspace{0.2cm}
			\includegraphics[width=0.22\textwidth]{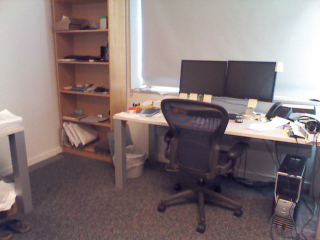}
			\includegraphics[width=0.22\textwidth]{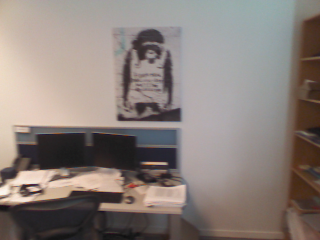}
			\includegraphics[width=0.22\textwidth]{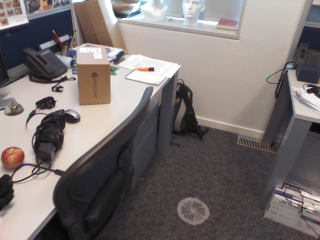}
			\includegraphics[width=0.22\textwidth]{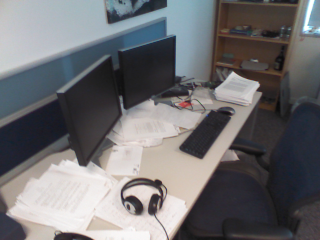}
			\begin{subfigure}[h]{0.22\textwidth}
				\includegraphics[width=\textwidth]{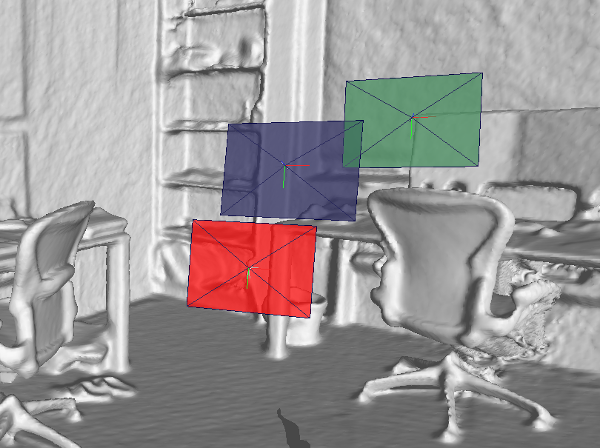}
				\subcaption{$10.8^\circ$-$8.1^\circ$, \SI{0.12}\meter -\SI{0.04}\meter}
			\end{subfigure}
			\begin{subfigure}[h]{0.22\textwidth}
				\includegraphics[width=\textwidth]{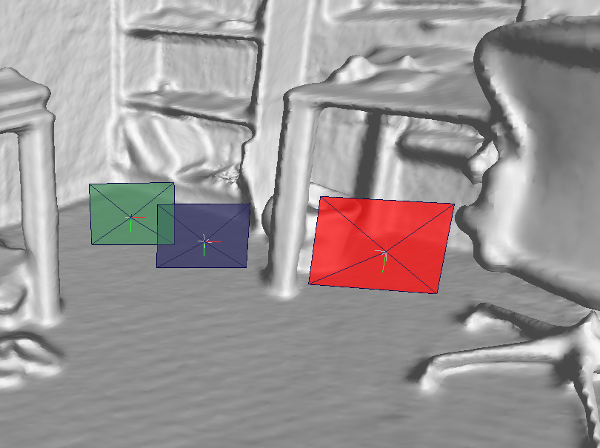}
				\subcaption{$9.8^\circ$-$6.2^\circ$, \SI{0.13}\meter -\SI{0.07}\meter}		
			\end{subfigure}
			\begin{subfigure}[h]{0.22\textwidth}
				\includegraphics[width=\textwidth]{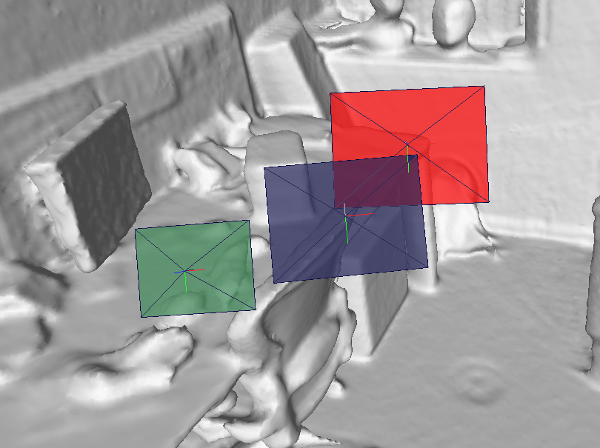}
				\subcaption{$14.2^\circ$-$12.1^\circ$, \SI{0.27}\meter -\SI{0.23}\meter}
			\end{subfigure}
			\begin{subfigure}[h]{0.22\textwidth}
				\includegraphics[width=\textwidth]{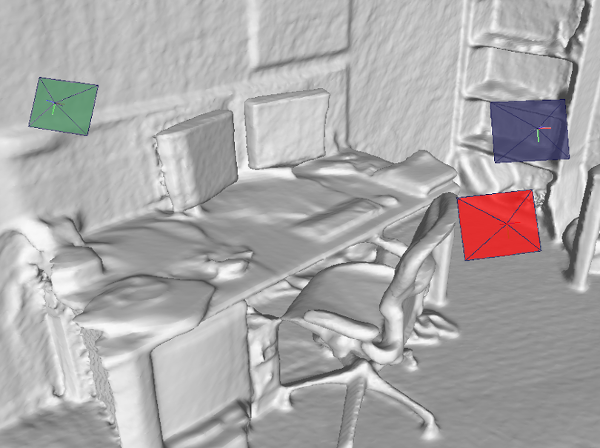}
				\subcaption{$30.9^\circ$-$29.8^\circ$, \SI{0.41}\meter -\SI{0.42}\meter}
			\end{subfigure}
			
		\end{center}
		\caption{RGB input images (first row) and the corresponding resulting camera poses (second row), visualized in a reconstruction of the given scene (Stairs, Red Kitchen, Office). For each frame the ground truth (\textcolor{green}{green}), initially regressed pose (\textcolor{red}{red}) and optimized pose using the proposed adversarial refinement (\textcolor{blue}{blue}) are displayed. Below each image initially regressed (left values) and refined (right values) rotation and translation errors are given.}
		\label{fig:qual1}
	\end{minipage}
\end{figure}
\pagebreak
\clearpage

\begin{figure}[h]
	\begin{minipage}[t]{\textwidth}			
		
		\captionsetup[subfigure]{labelformat=empty}
		\begin{center}
			\includegraphics[width=0.22\textwidth]{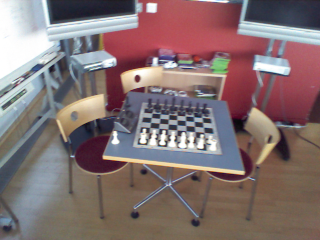}
			\includegraphics[width=0.22\textwidth]{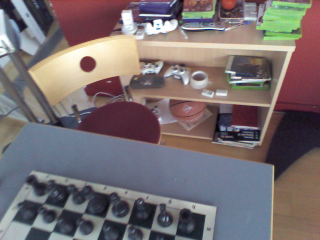}
			\includegraphics[width=0.22\textwidth]{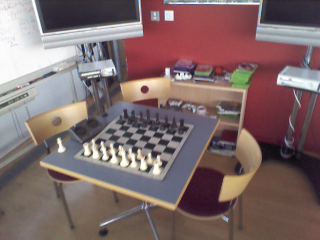}
			\includegraphics[width=0.22\textwidth]{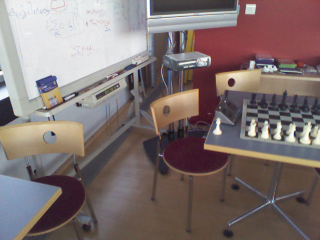}
			
			\begin{subfigure}[h]{0.22\textwidth}
				\includegraphics[width=\textwidth]{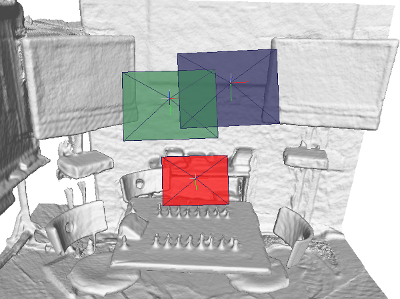}
				\subcaption{$4.9^\circ$-$3.6^\circ$, \SI{0.15}\meter -\SI{0.07}\meter}
			\end{subfigure}
			\begin{subfigure}[h]{0.22\textwidth}
				\includegraphics[width=\textwidth]{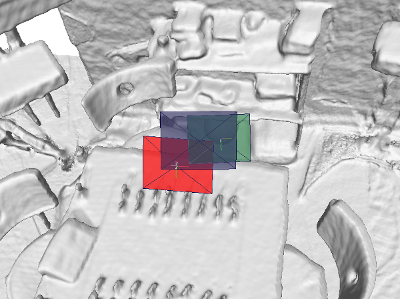}
				\subcaption{$3.2^\circ$-$1.4^\circ$, \SI{0.1}\meter -\SI{0.08}\meter}
			\end{subfigure}
			\begin{subfigure}[h]{0.22\textwidth}
				\includegraphics[width=\textwidth]{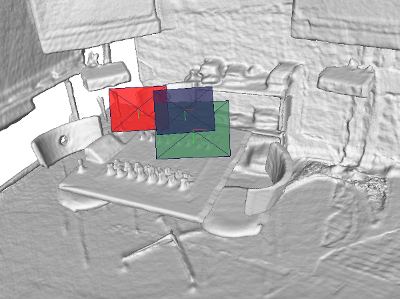}
				\subcaption{$2.8^\circ$-$1.7^\circ$, \SI{0.05}\meter -\SI{0.04}\meter}
			\end{subfigure}
			\begin{subfigure}[h]{0.22\textwidth}
				\includegraphics[width=\textwidth]{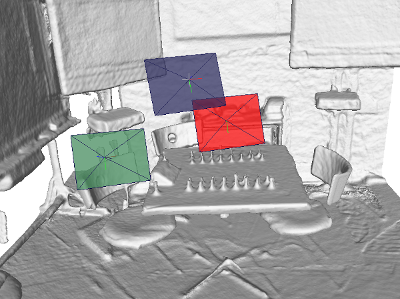}
				\subcaption{$1.3^\circ$-$2.2^\circ$, \SI{0.21}\meter -\SI{0.17}\meter}
			\end{subfigure}
			
			\vspace{0.2cm}
			\includegraphics[width=0.22\textwidth]{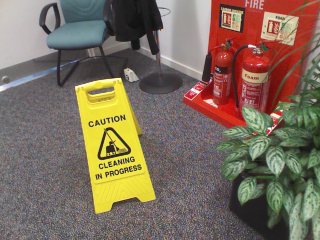}
			\includegraphics[width=0.22\textwidth]{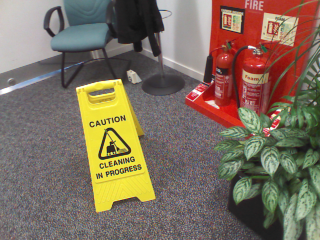}
			\includegraphics[width=0.22\textwidth]{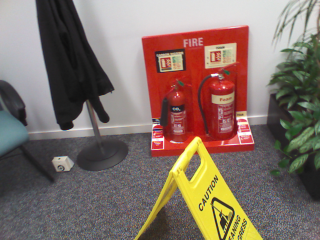}
			\includegraphics[width=0.22\textwidth]{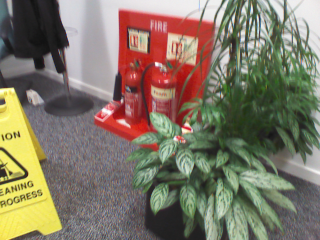}
			\begin{subfigure}[h]{0.22\textwidth}
				\includegraphics[width=\textwidth]{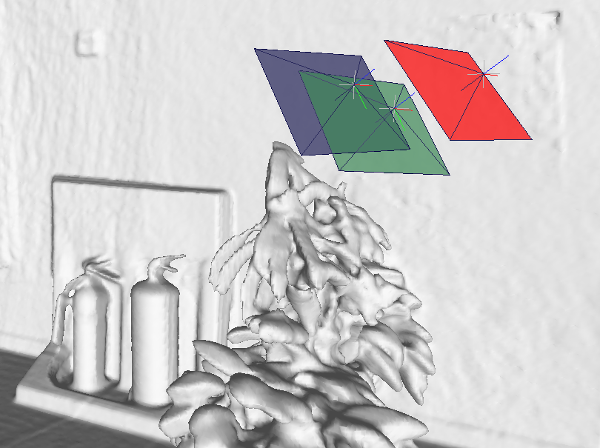}
				\subcaption{$5.9^\circ$-$1.9^\circ$, \SI{0.23}\meter -\SI{0.14}\meter}
			\end{subfigure}
			\begin{subfigure}[h]{0.22\textwidth}
				\includegraphics[width=\textwidth]{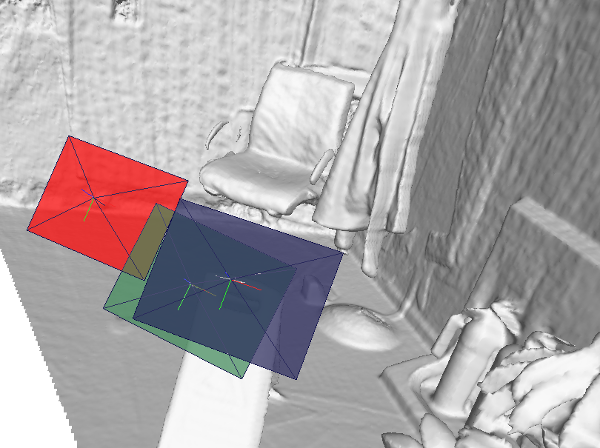}
				\subcaption{$7.8^\circ$-$6.4^\circ$, \SI{0.14}\meter -\SI{0.1}\meter}		
			\end{subfigure}
			\begin{subfigure}[h]{0.22\textwidth}
				\includegraphics[width=\textwidth]{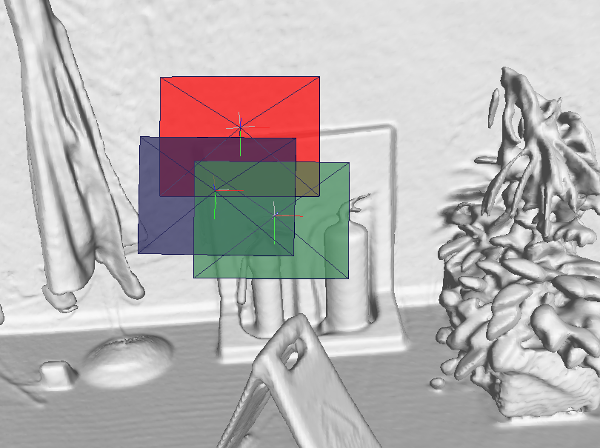}
				\subcaption{$4.5^\circ$-$2.8^\circ$, \SI{0.05}\meter -\SI{0.03}\meter}
			\end{subfigure}
			\begin{subfigure}[h]{0.22\textwidth}
				\includegraphics[width=\textwidth]{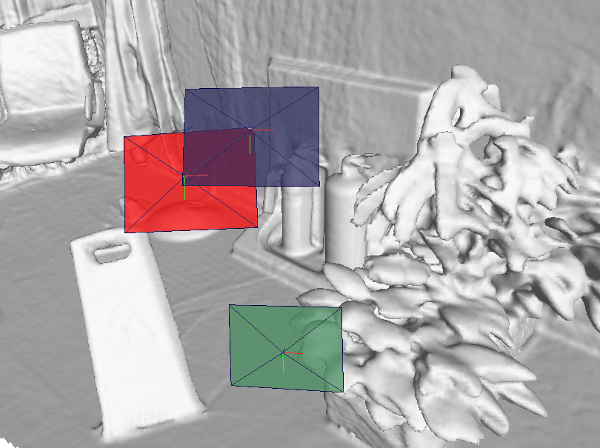}
				\subcaption{$6.7^\circ$-$8.0^\circ$, \SI{0.24}\meter -\SI{0.24}\meter}
			\end{subfigure}
			\hfill	
			
			\vspace{0.2cm}
			\includegraphics[width=0.22\textwidth]{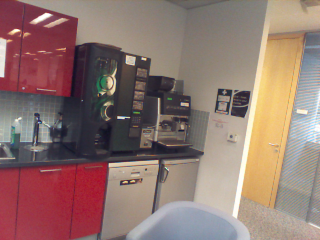}
			\includegraphics[width=0.22\textwidth]{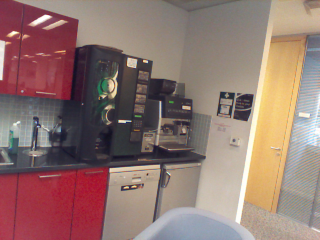}
			\includegraphics[width=0.22\textwidth]{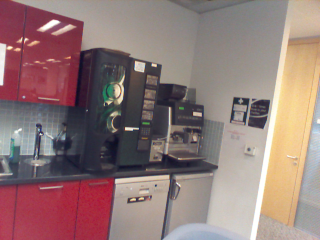}
			\includegraphics[width=0.22\textwidth]{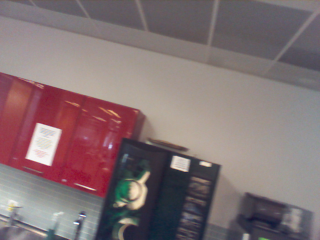}
			\begin{subfigure}[h]{0.22\textwidth}
				\includegraphics[width=\textwidth]{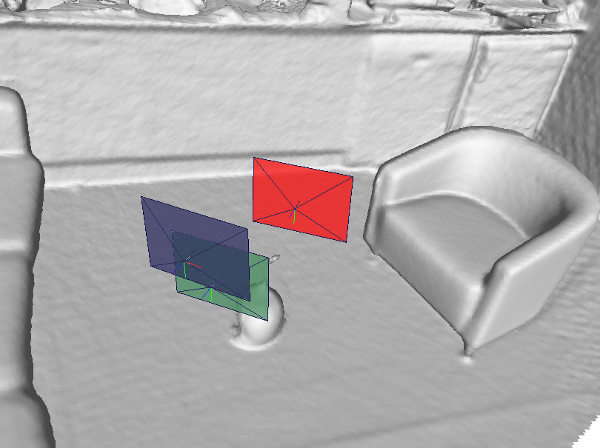}
				\subcaption{$5.1^\circ$-$3.0^\circ$, \SI{0.1}\meter -\SI{0.08}\meter}
			\end{subfigure}
			\begin{subfigure}[h]{0.22\textwidth}
				\includegraphics[width=\textwidth]{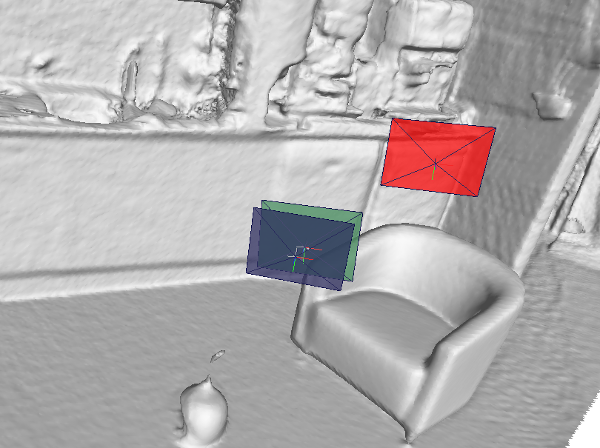}
				\subcaption{$3.3^\circ$-$1.4^\circ$, \SI{0.18}\meter -\SI{0.1}\meter}		
			\end{subfigure}
			\begin{subfigure}[h]{0.22\textwidth}
				\includegraphics[width=\textwidth]{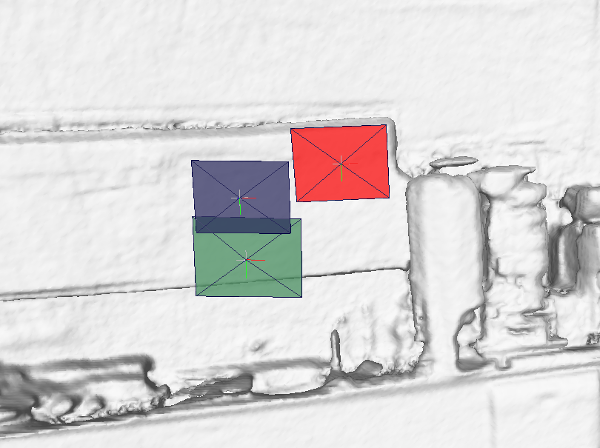}
				\subcaption{$2.7^\circ$-$1.9^\circ$, \SI{0.13}\meter -\SI{0.07}\meter}
			\end{subfigure}
			\begin{subfigure}[h]{0.22\textwidth}
				\includegraphics[width=\textwidth]{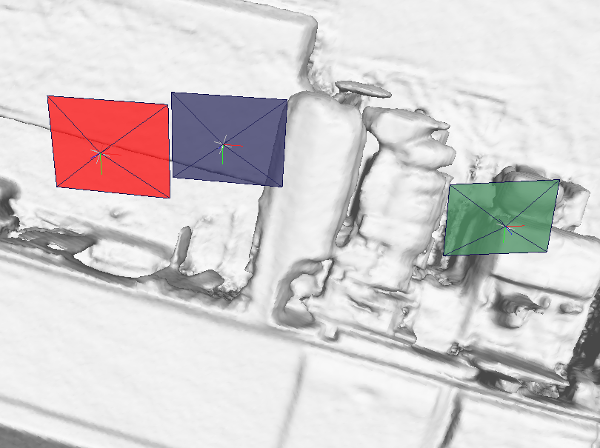}
				\subcaption{$9.7^\circ$-$10.0^\circ$, \SI{0.43}\meter -\SI{0.30}\meter}
			\end{subfigure}
			
		\end{center}
		\caption{RGB input images (first row) and the corresponding resulting camera poses (second row), visualized in a reconstruction of the given scene (Chess, Fire, Pumpkin). For each frame the ground truth (\textcolor{green}{green}), initially regressed pose (\textcolor{red}{red}) and optimized pose using the proposed adversarial refinement (\textcolor{blue}{blue}) are displayed. Below each image initially regressed (left values) and refined (right values) rotation and translation errors are given.}
		\label{fig:qual2}
	\end{minipage}
\end{figure}
\pagebreak
\clearpage
We show additional qualitative results of our method evaluated on the remaining scenes of the 7-Scenes dataset. RGB input images and the corresponding resulting camera poses, visualized in a reconstruction of the given scene, are shown in Figures \ref{fig:qual1} and \ref{fig:qual2}. For each frame the ground truth, initially regressed pose and optimized pose using the proposed pose refinement are displayed. Rotation and translation errors of the regressed and refined camera poses are shown in the caption of each image pair.
%\section{Statistics on the 7-Scenes dataset}
%In addition, we report more statistics on the 7-scenes dataset, which can be found in Table \ref{tab:stats}.

\section{Network Architectures}
Further, the network architectures used to train the models described in this paper are given in more detail. For simplicity, we abbreviate fully-connected layers as $FC$, convolutional layers as $C$ and average pooling layers as $AP$, where the resulting feature dimensionality is given as numbers after the respective layer. Further $ELU$ stand for the exponential linear unit, whereas $S$ describes the sigmoid function.\\

\textbf{Camera Pose Regression Network} The camera pose regression network consists of a ResNet-34 - $AP2048$ - $FC2048$ after which two fully connected layers for rotation $FC3/4$ and translation $FC3$ follow.\\

\textbf{Feature Extraction Network} This network consists of a pre-trained ResNet-18 - $GP512$ - $FC60/FC70$. All parameters of this network are fixed during training.\\

\textbf{Discriminator Network} The extracted features are concatenated with the replicated 6 or 7-dimensional pose vector and fed to the discriminator network, consisting of $C32$ - $ELU$ - $C16$ - $ELU$ - $C1$ - $S$.
\section{Runtime Evaluation}
Table \ref{tab:runtime} shows the computational times of the individual parts of our method evaluated for one frame. Pose refinement is calculated for 30 iterations. The method is implemented in Python and PyTorch and run on a 11GB NVIDIA GeForce RTX 2080 graphics card and 64 GB Intel Core i7.
\begin{table}[h]
	\caption{Computational times.}
	\resizebox{0.49\textwidth}{!}{
		\begin{tabular}{ccc|c}
			Pose regression & Feature extraction & Pose refinement & Overall\\
			4.5ms & 3ms & 42ms & $\sim$ 50ms \\
		\end{tabular}
	}
	\label{tab:runtime}
\end{table}
%\pagebreak
%\clearpage

{\small
\bibliographystyle{ieee}

}

\end{document}